# Near-infrared and visible-light periocular recognition with Gabor features using frequency-adaptive automatic eye detection

*Fernando Alonso-Fernandez, Josef Bigun*

*School of Information Science, Computer and Electrical Engineering, Halmstad University, Box 823, Halmstad SE 301-18, Sweden*

*E-mail: feralo@hh.se; josef.bigun@hh.se*

**Abstract:** Periocular recognition has gained attention recently due to demands of increased robustness of face or iris in less controlled scenarios. We present a new system for eye detection based on complex symmetry filters, which has the advantage of not needing training. Also, separability of the filters allows faster detection via one-dimensional convolutions. This system is used as input to a periocular algorithm based on retinotopic sampling grids and Gabor spectrum decomposition. The evaluation framework is composed of six databases acquired both with near-infrared and visible sensors. The experimental setup is complemented with four iris matchers, used for fusion experiments. The eye detection system presented shows very high accuracy with near-infrared data, and a reasonable good accuracy with one visible database. Regarding the periocular system, it exhibits great robustness to small errors in locating the eye centre, as well as to scale changes of the input image. The density of the sampling grid can also be reduced without sacrificing accuracy. Lastly, despite the poorer performance of the iris matchers with visible data, fusion with the periocular system can provide an improvement of more than 20%. The six databases used have been manually annotated, with the annotation made publicly available.

## 1 Introduction

Periocular recognition has gained attention recently in the biometrics field [1–3] with some pioneering works already in 2002 [4] (although authors here did not call the local eye area 'periocular'). Periocular refers to the face region in the immediate vicinity of the eye, including the eye, eyelids, lashes and eyebrows. Although face and irises have been extensively studied [5, 6], the periocular region has emerged as a promising trait for unconstrained biometrics, following demands for increased robustness of face or iris systems under less constrained conditions [1]. With a surprisingly high discrimination ability, this region can be easily obtained with existing setups for face and iris, and the requirement of user cooperation can be relaxed. Some studies also suggest that it may be as discriminative by itself as the face as a whole [7, 8]. Moreover, the periocular region appears in iris images, so fusion with the iris texture has potential to improve the overall recognition [9]. It has also another advantages, such as its availability over a wide range of distances even when the iris texture cannot be reliably obtained because of low resolution (high distances) or under partial face occlusion (close distances) [10].

In previous research [11], we proposed a periocular recognition system based on retinotopic sampling grids positioned in the pupil centre, followed by Gabor decomposition at different frequencies and orientations. This setup has been used previously in texture analysis [12], facial landmark detection and face recognition [4] and real-time face tracking and liveness assessment [13, 14], with high discriminative capabilities. The periocular system proposed, which we evaluated on two databases with near-infrared (NIR) illumination, achieved competitive verification rates in comparison with existing periocular approaches [1]. As in most studies on periocular recognition, it also relied on selecting manually the location of the periocular region. In a posterior study [15], we proposed a system for eye localisation based on complex symmetry filters. This system has the advantage of not needing training. Moreover, by using two-dimensional (2D) separable filters, detection can be done quite fastly by few 1D convolutions. The proposed detection system was able to accurately detect the centre of the eye when using good quality iris data acquired with close-up NIR cameras, and it still worked good with most difficult data acquired with a webcam. We also used this detection system as input of our periocular recognition system [15]. Results showed that the periocular system is quite robust to inaccuracies in detecting the centre of the eye. We also carried out experiments with an iris texture matcher based on 1D log-Gabor (LG) filters. Despite the poor performance of this iris matcher with webcam images, we observed that the fusion with our periocular system resulted in an improved performance.

### 1.1 Contributions

This paper extends our two previous studies [11, 15] with new developments and experiments. The eye detection and periocular recognition systems are described more formally and in greater detail. We make use of a more comprehensive set of data coming from six different databases, in contraposition to our previous works, which only made use of two databases. We also add three new iris matchers to our experimental framework. The contribution of this paper is therefore multi-fold:

1. *Eye localisation system based on symmetry filters:* The eye detection system has been improved with the addition of two steps. The first one is concerned with frequency estimation of the iris image [16], which in this case is referred to the average transition width of the image edges. Our eye detection system relies on the computation of the orientation field of the input image, which is implemented via convolutions with sampled Gaussian-derivative filters that resemble the ordinary gradient in 2D [17]. In our previous work [11], the size of the derivative filters (which corresponds to the width of the transitions that we want to detect) was set to a fixed value. With the new developments of this project, this parameter is fixed dynamically, based on the analysis of the input image. Reported results show that this results in an increased accuracy of our eye detection system. We also add an eyelash removal step via rank filters [18]. Eyelashes appear as prominent vertical edges in the orientation field, which may mislead the symmetry filter used for eye localisation. A novelty with respect to the system proposed in [18] is that the size of the rank filter is also adjusted dynamically according to the estimated image frequency.

2. *Periocular recognition system based on sampling grids and Gabor spectrum decomposition:* The eye detection system is validated by using the detected eye centre as input to our periocular recognition system. Two different sampling grid configurations are evaluated within the periocular recognition system, with dense and coarse sampling. On the basis of our experiments, the density of the grid can be reduced without sacrificing too much accuracy, allowing computational savings in the feature extraction process. Although not directly comparable because of the use of different databases for the experiments, our system achieves competitive verification rates in comparison with existing periocular recognition approaches [1]. It is also shown that, when eye detection is done with sufficient accuracy, the recognition performance does not suffer a significant drop. This demonstrates that the periocular recognition system can tolerate small deviations in the position of the sampling grid. These results are in line with our previous observations [11, 15], that are validated here with a bigger test set. Lastly, a novel observation in this paper is that the periocular system is able to cope with certain degree of variation in the scale (size) of the eye.

3. *Fusion of periocular and iris information:* We evaluate four iris matchers based on 1D LG filters [19], local intensity variations in iris textures proposed by Christian Rathgeb *et al.* (CR) [20], discrete-cosine transform (DCT) [21] and cumulative-sum-based grey change analysis proposed by Ko *et al.* (KO) [22], which are also fused with the periocular matcher. With NIR images, the iris matchers are, in general, considerably better in our experiments. On the other hand, they show a poorer performance with challenging images with VW illumination, but the fusion of iris and periocular systems can result in a considerable improvement. Despite the adverse acquisition conditions of the VW databases and the lower image resolution, it is worth noting that the (even smaller) iris texture is still able to provide complementary identity information to the periocular system.

4. *Evaluation on multiple databases captured both with NIR and visible (VW) illumination*: In our previous papers [11, 15], we used only two databases on each case. Here, we use six different databases: BioSec baseline [23], CASIA-Iris Interval v3 of the Institute of Automation, Chinese Academy of Sciences (CASIA) [24], IIT Delhi v1.0 (IITD v1.0 database) of the Indian Institute of Technology (IIT) [25], MobBIO [26], UBIRIS v2 [27] and Notre Dame ND-IRIS-0405 [28]. Five of them have been captured with NIR sensors, whereas the other two, with sensors in VW range. These provide a comprehensive and heterogeneous test set, allowing extensive validation experiments of our developments.

5. *Groundtruth database of iris segmentation data*: The databases used in this paper have been annotated manually by an operator, enabling accurate evaluation of the eye localisation system. The annotation is also used as input of the periocular recognition system, allowing to test its robustness against inaccuracies in the detection of the eye centre given by small errors of the eye detection system. The iris segmentation groundtruth has been made available to the research community (http://www.islab.hh.se/mediawiki/index.php/Iris_Segmentation_Groundtruth and http://www.wavelab.at/sources) under the name of Iris Segmentation (IRISSEG) Database [29].

### 1.2 Literature review

#### 1.2.1 Eye detection and iris segmentation:

Most studies of periocular recognition have not focused on detection of the periocular region (it is manually extracted), but on feature extraction only. Only Park *et al.* [2] used a Viola–Jones face detector [30] plus heuristics measurements (not specified) to extract the periocular region, so successful extraction relied on an accurate detection of the whole face. Traditional iris segmentation techniques based on edge information, such as the integro-differential operator [31] or the Hough transform [32], may not be reliable under challenging conditions either. Jillela *et al.* [10] evaluated these two iris segmentation approaches, together with newer approaches designed to handle challenging iris images, such as geodesic active contours [33], active contours without edges [34] or directional ray detection [35]. They used the difficult face and ocular challenge series (FOCS) database of periocular images, which were captured from subjects walking through a portal with NIR illumination in an unconstrained environment. With the traditional segmentation approaches [31, 32], segmentation accuracy was below 55%, whereas the other approaches were in the range 85–90%. The latter methods, however, are much more computationally expensive, as noted in [10].

Many iris segmentation techniques include rough location of dark pixels of the pupil by image thresholding [36]. However, this may not work in the presence of non-uniform illumination or other adverse conditions where the periocular modality is precisely intended to achieve its highest potential. Some works deal with the issue of locating the eye position without relying on a full-face detector, with a summary of them given in

**Table 1** Overview of existing automatic eye detection works

| Approach | Features | Training | Database | Best accuracy |
|---|---|---|---|---|
| Smeraldi and Bigün [4] | Gabor filters | M2VTS (202 VW images) | M2VTS (349 VW images) XM2VTS (2388 VW images) | 99.3% (M2VTS) 99% (XM2VTS) |
| Uhl and Wild [37] | Viola–Jones detector of face sub-parts (OpenCV) | yes (n/a) | CASIA distance v4 (282 NIR images) Yale-B (252 VW images) | 96.4% (NIR) 99.2% (VW) |
| Jillela *et al.* [10] | correlation filter | 1000 eye images | FOCS (404 NIR images) | 95% |
| our approach | symmetry filters | no | six iris datasets: four NIR, two VW (6932 NIR images, 3050 VW) | 96% (NIR) 27% (VW) |

Table 1. The work by Smeraldi *et al.* [13] made use of sampling grids and Gabor features in a similar manner that our periocular recognition system, but for eye detection and face tracking purposes. More recently, in [37], Uhl and Wild used the OpenCV implementation of Viola–Jones detectors of face sub-parts [30]. An accuracy of 96.4/99.2% in the combined detection of face parts was reported using NIR/VW face images, respectively. Eye detection can be also a decisive preprocessing task to ensure successful segmentation of the iris texture in difficult images. In [10], Jillela *et al.* used a correlation filter [38] to detect the eye centre in their experiments with iris segmentation algorithms mentioned above, achieving a 95% success rate. Despite this good result in indicating the approximate position of the eye, the accuracy of iris segmentation algorithms in the challenging FOCS database was between 51 and 90%, as indicated above.

*1.2.2 Features for periocular recognition:* An overview of existing approaches for periocular recognition is given in Table 2. A recent review article has also been published in [1]. The most widely used approaches include local binary patterns (LBPs) [47] and, to a lesser extent, histogram of oriented gradients (HOGs) [48] and scale-invariant feature transform (SIFT) keypoints [49]. The use of different experimental setups and databases make difficult a direct comparison between existing works. The study of Park *et al.* [2] compares LBP, HOG and SIFT using the same data, with SIFT giving the best performance (rank-one recognition accuracy: 79.49%, equal error rate

**Table 2** Overview of existing periocular recognition works

| Approach | Features | Test database | Best accuracy (single eye) | | |
|---|---|---|---|---|---|
| | | | | EER | Rank-one |
| Smeraldi and Bigün [4] | Gabor filters | M2VTS: Gabor (349 VW images) | M2VTS: Gabor | 0.3% | n/a |
| Park *et al.* [2] | HOG, LBP, SIFT[a] | FRGC v2.0 (1704 VW images) | FRGC: HOG | 21.78% | 66.64% |
| | | | FRGC: LBP | 19.26% | 72.45% |
| | | | FRGC: SIFT | 6.96% | 79.49% |
| Miller *et al.* [39] | LBP | FRGC (1230 VW images) | FRGC: LBP | 0.09% | 89.76% |
| | | FERET (162 VW images) | FERET: LBP | 0.23% | 74.07% |
| Adams *et al.* [40] | GEFE + LBP | FRGC (820 VW images) | FRGC: GEFE + LBP | n/a | 86.85% |
| | | FERET (108 VW images) | FERET: GEFE + LBP | n/a | 80.8% |
| Juefei-Xu *et al.* [41, 42] | LBP, WLBP, SIFT, DCT | FRGC (16 028 VW images) | FRGC: LBP + DWT | n/a | 53.2% |
| | Walsh masks, DWT, SURF | FG-NET (1002 VW images) | FRGC: LBP + DCT | n/a | 53.1% |
| | law masks, force fields | | FRGC: LBP + Walsh | n/a | 52.9% |
| | Gabor filters, LoG | | FG-NET: WLBP | 0.6% | 100% |
| Bharadwaj *et al.* [43] | ULBP, GIST | UBIRIS v2 (7409 VW images) | UBIRIS: ULBP | n/a | 54.30% |
| | | | UBIRIS: GIST | n/a | 63.34% |
| | | | UBIRIS: ULBP + GIST | n/a | 73.65% |
| Woodard *et al.* [8] | RG colour histogram | FRGC (4100 VW images) | FRGC: RG | n/a | 96.1% |
| | LBP | MBGC (911 NIR images) | FRGC: LBP | n/a | 95.6% |
| | | | FRGC: RG + LBP | n/a | 96.8% |
| | | | MBGC: LBP | n/a | 87% |
| Woodard *et al.* [9] | LBP | MBGC (1052 NIR images) | MBGC: LBP | 0.21% | 92.5% |
| Padole and Proenca [44] | HOG, LBP, SIFT | UBIPr (10 950 VW images) | UBIPr: HOG + LBP + SIFT | n/a | ~20% |
| Hollingsworth *et al.* [45] | human observers | NIR (120 subjects) | NIR: human | n/a | 92% |
| Hollingsworth *et al.* [3] | human observers | VW (210 subjects) | VW: human | n/a | 88.4% |
| | | NIR (210 subjects) | NIR: human | n/a | 78.8% |
| Mikaelyan *et al.* [46] | symmetry patterns (SAFE) | BioSec (1200 NIR images) | BioSec: SAFE | 12.81% | n/a |
| | | MobBIO (800 VW images) | MobBIO: SAFE | 11.96% | n/a |
| our approach | Gabor features | BioSec (1200 NIR images) | BioSec: Gabor | 10.56% | 66%[b] |
| | | CASIA Interval v3 (2655 NIR images) | CASIA: Gabor | 14.53% | n/a |
| | | IIT Delhi v1.0 (2240 NIR images) | IITD: Gabor | 2.5% | n/a |
| | | MobBIO (800 VW images) | MobBIO: Gabor | 12.32% | 75%[b] |
| | | UBIRIS v2 (2250 VW images) | UBIRIS: Gabor | 24.4% | n/a |

[a]The acronyms recorded in this table are defined fully in the referenced papers.
[b]Results reported in [15].

(EER): 6.95%), followed by LBP (rank-one: 72.45%, EER: 19.26%) and HOG (rank-one: 66.64%, EER: 21.78%). Other works with LBPs, however, report rank-one accuracies above 90% and EER rates below 1% [8, 9, 39]. Gabor features were also proposed in a seminal work of 2002 [4], which have served as inspiration for our periocular system, although this work did not call the local eye area 'periocular'. Here, the authors used three machine experts to process Gabor features extracted from the facial regions surrounding the eyes and the mouth, achieving very low error rates (EER $\leq$ 0.3%). We also have recently proposed a new system based on detection of local symmetry patterns (which we call SAFE features) [46], with reported EER rates of ~12%. Lastly, in the extensive experiments of this paper with our Gabor-based periocular system, we report EER rates of 2.5–14.53% (NIR data) and 12.32–24.4% (VW data); and rank-one accuracies of 66% (NIR data) and 75% (VW data) [15]. Another important set of research works have concentrated their efforts in the fusion of different algorithms. For example, Bharadwaj *et al.* [43] fused uniform LBPs (ULBP) with a global descriptor (GIST) consisting of perceptual dimensions related with scene description (image naturalness, openness, roughness, expansion and ruggedness). The best result, obtained by the fusion of both systems, was a rank-one accuracy of 73.65%. Juefei-Xu *et al.* [41, 42] fused LBP and SIFT with other local and global feature extractors including Walsh masks [50], Law's Masks [51], DCT [52], DWT [53], Force Fields [54], speeded-up robust feature (SURF) [55], Gabor filters [56] and Laplacian of Gaussian. The best result obtained was a rank-one accuracy of 53.2% by fusion of DWT and LBP. Finally, Hollingsworth *et al.* [3] evaluated the ability of (untrained) human observers to compare pairs of periocular images, resulting in a rank-one accuracy of 88.4% (VW data) and 78.8% (NIR data).

Comparison of periocular with face or iris is also done in some cases. For example, Park *et al.* [2] reported a rank-one accuracy of 99.77% using the whole face, but when the full face is not available (simulated by synthetically masking the face below the nose region), accuracy fell to 39.55%. This points out the strength of periocular recognition when only partial face images are available, for example, in criminal scenarios with surveillance cameras, where it is likely that the perpetrator masks parts of his face. In the same direction, Miller *et al.* [7] found that at extreme values of blur or down-sampling, periocular recognition performed significantly better than face. On the other hand, both face and periocular matching using LBPs under uncontrolled lighting were very poor, indicating that LBPs are not well suited for this scenario. Finally, Woodard *et al.* [9] fused periocular and iris information from NIR portal data. Using a traditional iris algorithm based on Gabor filters [31], they found that periocular identification performed better than iris, and the fusion of the two modalities performed best. In most of these studies, periocular images were acquired in the VW range (see third row of Table 2). Hollingsworth *et al.* [3] compared the use of visible-light and NIR light images in periocular and iris recognition. According to this study, periocular on VW images works better than on NIR because visible-light images show melanin-related differences that do not appear in NIR images. This is supported by other studies which use VW and NIR data simultaneously in the experiments [8], but this is not the case in present paper, with our periocular system achieving better performance on some NIR periocular databases than on the VW ones. On the other hand, many iris systems work with NIR illumination because of higher reflectivity of the iris tissue in this range [31]. Unfortunately, the use of more relaxed scenarios will make NIR light unfeasible (e.g. distant acquisition, mobile devices etc.) so there is a high pressure in the research field to the development of iris algorithms capable of working with visible light [57].

### 1.3 Paper organisation

This paper is organised as follows. Section 2 describes our eye detection system. The periocular recognition system and iris matcher used in this paper are described in Sections 3 and 4, respectively. Section 5 describes the evaluation framework, including the databases and protocol used. Results obtained are presented in Sections 6 and 7, followed by conclusions in Section 8.

## 2 Eye localisation

We propose the use of symmetry features for eye localisation. Symmetry features enable the description of symmetric patterns such as lines, circles, parabolas and so on (Fig. 1). These features are extracted via symmetry filters, (1), which output how much of a certain symmetry exists in a local image neighbourhood [17, 58]. An overview of the system proposed for eye localisation is shown in Fig. 2.

### 2.1 Symmetry filters

Symmetry filters are a family of filters computed from symmetry derivatives of Gaussians. The $n$th symmetry derivative of a Gaussian, $\Gamma^{\{n,\sigma^2\}}$, is obtained by applying the partial derivative operator $D_x + iD_y = (\partial/\partial x) + i(\partial/\partial y)$, called first symmetry derivative, to a Gaussian

$$\Gamma^{\{n,\sigma^2\}} = \begin{cases} (D_x + iD_y)^n g(x, y), & (n \geq 0) \\ (D_x - iD_y)^{|n|} g(x, y), & (n < 0) \end{cases} \tag{1}$$

Since $D_x + iD_y$ and $(-1/\sigma^2)(x + iy)$ behave identically when

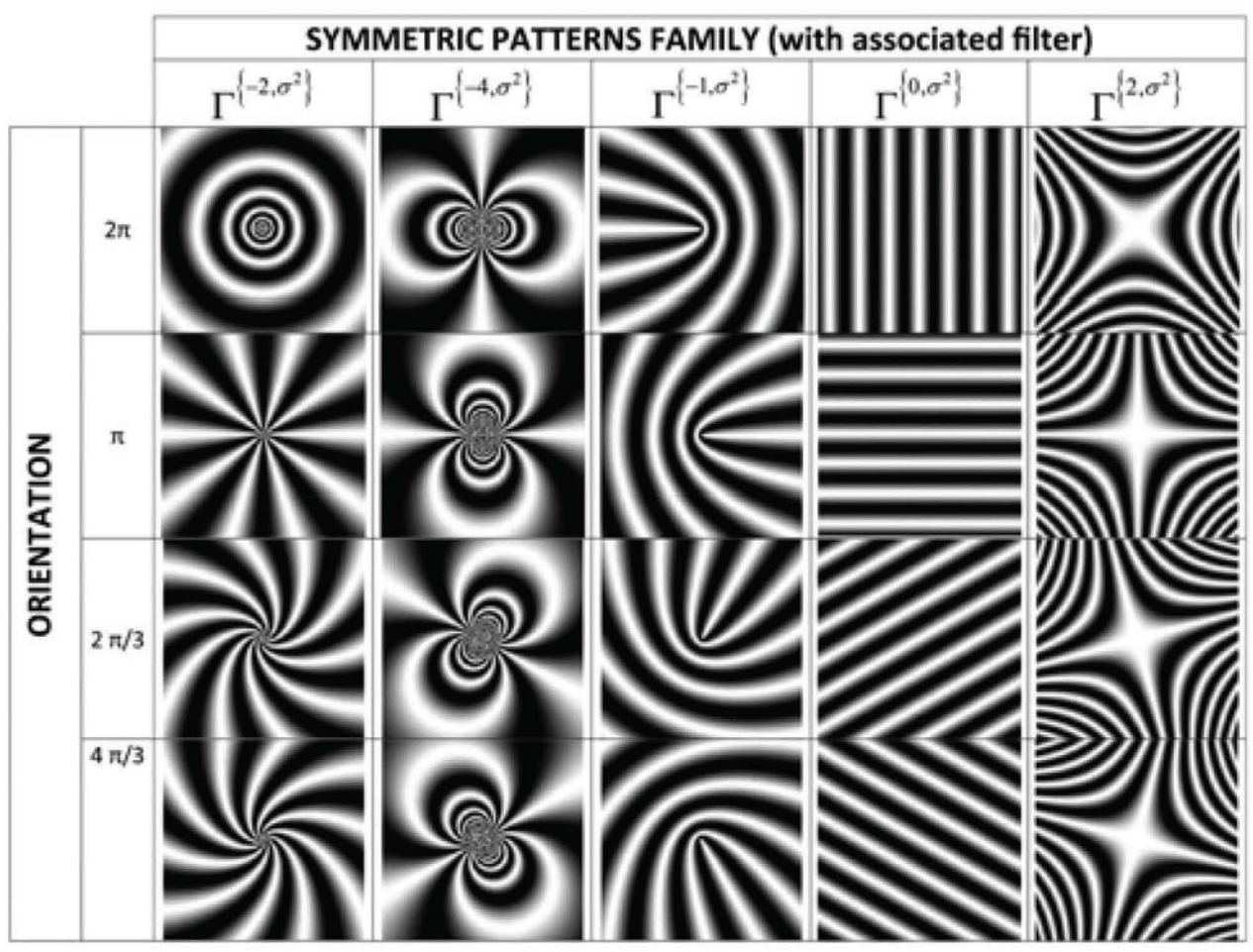

**Fig. 1** *Example of symmetric patterns*

Each column represents one family of patterns differing only by their orientation (given in column 2). The associated filter suitable to detect each family (2) is also indicated in row 2

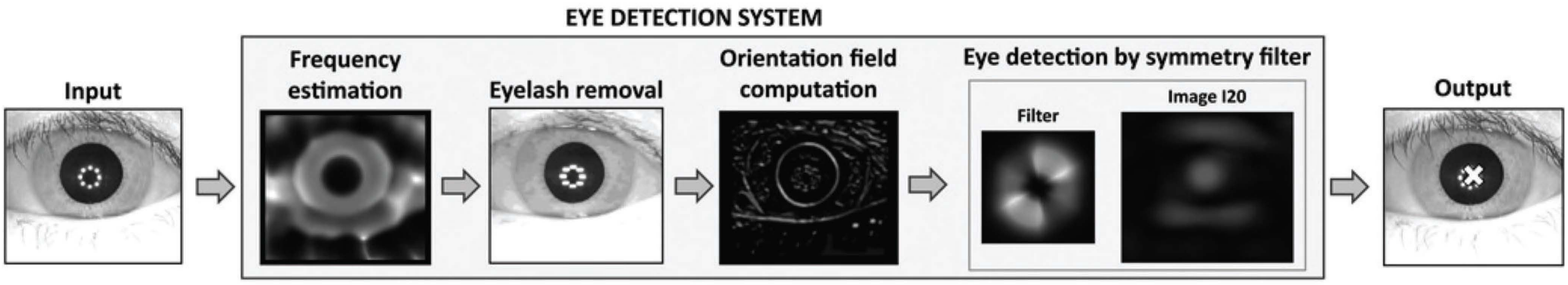

**Fig. 2** *Eye detection using symmetry filters*

Hue in colour images encodes the direction, and the saturation represents the complex magnitude. To depict the magnitude, they are re-scaled, so the maximum saturation represents the maximum magnitude, while black represents zero magnitude. Zero angle is given by red colour. It can be observed that $I_{20}$ shows a prominent region in red colour around the neighbourhood of the eye centre. The detected eye centre is marked in the original iris image too. Iris image is from CASIA-Iris Interval v3 database

acting on a Gaussian [17, 58], (1) can be rewritten as

$$\Gamma^{\{n,\sigma^2\}} = \begin{cases} \left(-\dfrac{1}{\sigma^2}\right)^n (x+\mathrm{i}y)^n g(x, y), & (n \geq 0) \\ \left(-\dfrac{1}{\sigma^2}\right)^{|n|} (x-\mathrm{i}y)^{|n|} g(x, y), & (n < 0) \end{cases} \quad (2)$$

The interest is that these symmetry derivatives of Gaussians are able to detect patterns as those of Fig. 1 through the computation of the second-order complex moment of the power spectrum via [17]

$$I_{20} = \left\langle \Gamma^{\{n,\sigma_2^2\}}, h \right\rangle \quad (3)$$

where $h$ is the complex-valued orientation tensor field given by

$$h = \left\langle \Gamma^{\{1,\sigma_1^2\}}, f \right\rangle^2 \quad (4)$$

and $f$ is the image under analysis [17]. Parameter $\sigma_1$ defines the size of the derivation filters used in the computation of image $h$, whereas $\sigma_2$, used in the computation of $I_{20}$, defines the size extension of the sought pattern.

For each family of symmetric patterns, there is a symmetry filter $\Gamma^{\{n,\sigma^2\}}$ (indexed by $n$) suitable to detect the whole family [59]. Fig. 1 indicates the filters that are used to detect each family. The local maxima in $|I_{20}|$ gives the location, whereas the argument of $I_{20}$ at maxima locations gives the group orientation of the detected pattern (except for the first family in Fig. 1, $n = -2$, where the 'orientation' represents the chirality of the curves). Therefore, $I_{20}$ encodes how much of a certain type of symmetry exists in a local neighbourhood of the image $f$. In addition, a single symmetry filter $\Gamma^{\{n,\sigma^2\}}$ is used for the recognition of the entire family of patterns, regardless of their orientation (or chirality). Symmetry filters have been successfully applied to a wide range of detection tasks such as cross-markers in vehicle crash tests [60], core-points and minutiae in fingerprints [61, 62] or iris boundaries [63]. The beauty of this method is even more emphasised by the fact that $I_{20}$ is computed by filtering in Cartesian coordinates without the need of transformation to curvilinear coordinates (which is implicitly encoded in the filter).

### 2.2 Eye detection process

We use the filter of order $n = -2$ to detect the eye position in a given image. By assuming that iris boundaries can be approximated as circles, the eye can be detected with the pattern of concentric circles shown in Fig. 1 (top left). Despite the inner (pupil) and outer (sclera) boundaries of the iris are not concentric, we exploit the evidence that the pupil is fully contained within the sclera boundary, with the centre of both circles in close vicinity [31]. Owing to the separable property of 2D Gaussians, the filter can be re-written as

$$\begin{aligned} \Gamma^{\{-2,\sigma^2\}} &= \left(-\frac{1}{\sigma^2}\right)^2 (x-\mathrm{i}y)^2 g(x)g(y) \\ &= \left(-\frac{1}{\sigma^2}\right)^2 (x^2 g(x)g(y) - y^2 g(y)g(x) - \mathrm{i}2xg(x)yg(y)) \end{aligned} \quad (5)$$

so the 2D convolutions can be computed by several 1D convolutions, achieving a considerable higher speed. Moreover, in computing $h$, 1D convolutions can be used, since

$$\begin{aligned} \Gamma^{\{1,\sigma^2\}} &= \left(-\frac{1}{\sigma^2}\right)^2 (x+\mathrm{i}y) g(x)g(y) \\ &= \left(-\frac{1}{\sigma^2}\right)^2 (xg(x)g(y) + \mathrm{i}yg(y)g(x)) \end{aligned} \quad (6)$$

After the computation of $I_{20}$, we search for local maxima in $|I_{20}|$ with a window of size $7 \times 7$. The maximum with highest magnitude is selected as the centre of the eye. Evidence of the pattern of concentric circles is given by an argument of the complex filter response equal to zero ($2\pi$ in Fig. 1). Thus, only maxima with absolute angle below a certain threshold are considered. An example of the detection process can be seen in Fig. 2.

### 2.3 Image frequency estimation

Computation of the orientation field via (4) is achieved by convolution of the input image $f$ with the first-order Gaussian-derivative filter

$$\Gamma^{\{1,\sigma^2\}} = (D_x + \mathrm{i}D_y) g(x, y) = \left(-\frac{1}{\sigma^2}\right)^2 (x+\mathrm{i}y) g(x, y) \quad (7)$$

This symmetry derivative filter resembles the ordinary gradient in 2D [17]. Parameter $\sigma$ controls the size of the

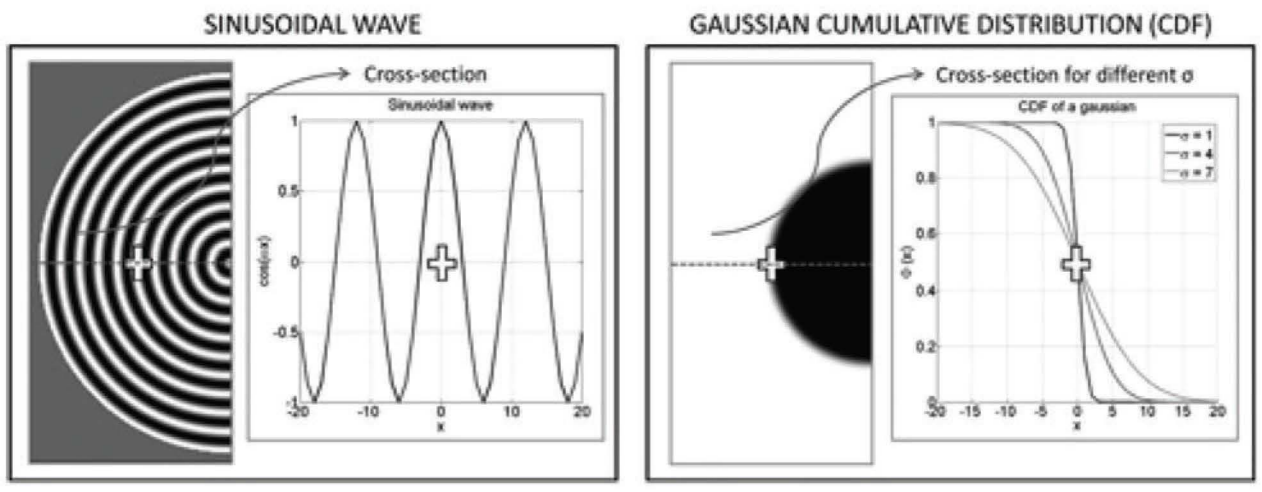

**Fig. 3** *Left: planar 2D sinusoid with absolute frequency $\omega_o = 2\pi/12$*

*Right:* 2D Gaussian CDF with transition of width $T = 24$ ($\sigma = 24/6 = 4$). On the right part of each subplot, it is shown the cross-section across the horizontal red line. For the CDF, cross-sections with different values of $\sigma$ are also shown derivation filters in the computation of the image $h$. This size, in turn, is related with the width of the edge transition that the filter will be able to detect (Fig. 3, right). We give in Fig. 4 an example of the orientation field obtained from an iris image with different $\sigma$s. As can be observed, using a low value of $\sigma$ results in parts of the sclera edge undetected. It also results in too much noise in the image, since the derivative filter is sensitive to edges of small width, which typically are the product of quick noisy oscillations. On the contrary, using a high value of $\sigma$ has the undesirable effect of making the detected edges too wide, which may merge some edges which are close to each other. This can be appreciated in the pattern of light reflections within the pupil, which become blurred.

In our previous work [15], the value of $\sigma$ was set heuristically to a fixed value. In this paper, we propose a method to estimate the average width of the transitions found in the edges of the image, which will be used to set $\sigma$. This will allow to customise the size of the derivative filters to each individual image. For this purpose, we make use of the method proposed in [16] to compute the image frequency map. This method is tailored to compute the local frequency of 2D sinusoidal waves (Fig. 3, left). Here, we model the target edge transitions in iris images as a cumulative distribution function (CDF) of a Gaussian, which in 1D is given by (Fig. 3, right)

$$\Phi(x) = 0.5\left[1 + \text{erf}\left(\frac{x}{\sqrt{2\sigma^2}}\right)\right] \tag{8}$$

It can be seen in Fig. 3 that the width $T$ of the CDF transition corresponds to approximately $3 \times \sigma$ to each side of the origin. Therefore, we define

$$T = 6 \times \sigma \tag{9}$$

The algorithm of [16] is used to estimate the width of the CDF transition as follows. We provide as input a set of 2D CDFs as those shown in Fig. 3, right, with different transition widths ($T_1$, $T_2$, …, $T_N$), for which we obtain an estimated frequency value ($F_1$, $F_2$, …, $F_N$) in the point marked with '+'. The correspondence between the output and the input parameters is then found by training a second-order polynomial: $T'(F) = aF^2 + bF + c$. We allow the parameter $T_i$ to vary between 2 and 22, which is a reasonable range for edge transitions in an iris image. Given an input iris image, we then apply this algorithm pixel-wise, see an example in Fig. 2. A global edge transition value of the whole image is finally computed by averaging the value of each pixel.

### 2.4 Eyelash removal

Eyelashes appear as prominent vertical edges in the orientation field, which may mislead the symmetry filter used for eye localisation. We incorporate an eyelash removal step based on the method proposed in [18]. For this purpose, prior to computation of the orientation field (4), a 1D rank filter is applied to the image. A rank filter is a filter whose response is based on ranking (ordering) the pixels contained in the image area encompassed by the filter. A 1D rank-$p$ filter of length $1 \times L$ will replace the centre pixel by the $p$th grey level in the $L$-neighbourhood of that pixel. After applying this filter, most eyelashes will be weakened or even eliminated (Fig. 2). A value of $L = 7$ and $p = 2$ is proposed in [18], which will be used in this paper. In addition, we propose to use a value of $L$ proportional to the estimated image frequency of the previous step.

## 3 Periocular recognition system

For periocular recognition, we use the system proposed in [11], which is based on the face detection and recognition system of [4, 13]. Input images are analysed with a retinotopic sampling sensor, whose receptive fields consist in a set of modified Gabor filters designed in the log-polar frequency plane. The system is described next.

### 3.1 Sampling grid

Our periocular system makes use of a sparse retinotopic sampling grid positioned in the eye centre. The grid has rectangular geometry, with sampling points distributed uniformly (Fig. 5). At each point, a Gabor decomposition of the image is performed, see Section 3.2. The sparseness of the sampling grid allows direct Gabor filtering in the image domain without needing the Fourier transform, with significant computational savings [4] and even feasibility in real time [14]. We evaluate two different grid configurations (see Fig. 5 and Table 3), one with a dense sampling, and another with a coarse sampling. Parameter $d_i$ indicates the distance between adjacent sampling points. This distance is constant for all images of the same database, so dimensions

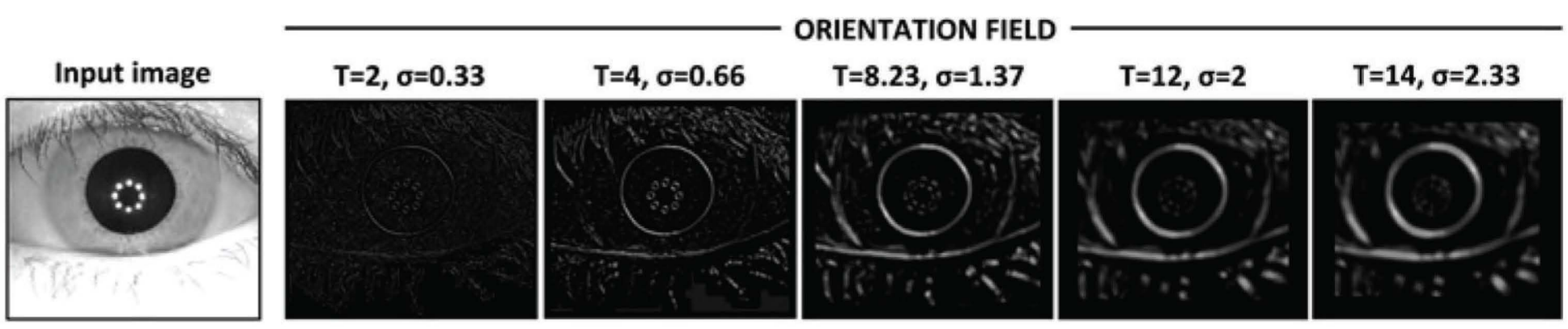

**Fig. 4** *Computation of the complex orientation field with different sizes of the derivation filters*

$T = 8.23$ is the average width of the image edges found with the algorithm of Section 2.3

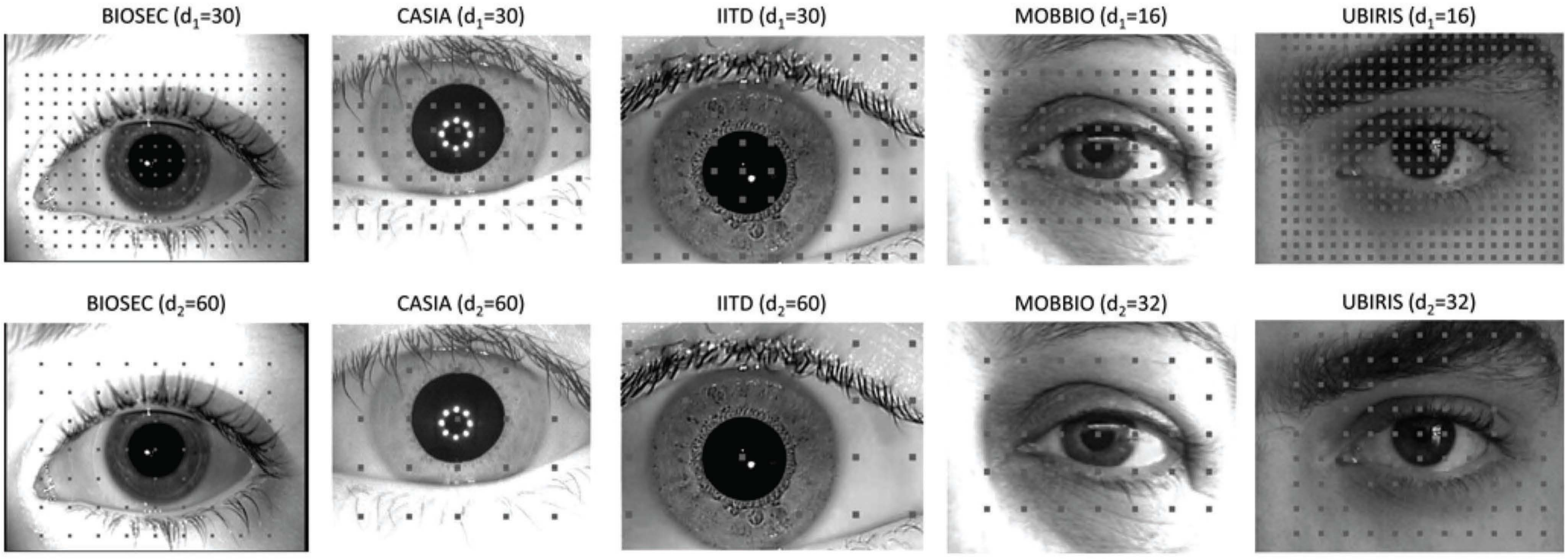

**Fig. 5** *Sampling grid showing different configurations with the databases used in this paper (images are resized to the same height)*

of the grid are not adapted to the scale of the input eye image. Owing to different image size (Table 4), dimensions of the sampling grid are set accordingly for each database. Images in UBIRIS show the eyebrows in many cases, so the grid has been made large in both vertical and horizontal directions to capture such region. With the other databases, the grid has been designed in such way that the eyelids and eyelashes (vertical direction) and the eyelids corners (horizontal direction) are sufficiently covered, whenever possible. The latter cannot be achieved with CASIA and IITD however, since eyelids corners fall outside the image limits in most cases.

### 3.2 Gabor decomposition of the power spectrum

The local power spectrum of the image is sampled at each cell of the grid by a set of modified Gabor filters organised in 5 frequency channels and 6 equally spaced orientation channels. For a filter tuned to orientation $\varphi_0$ and angular frequency $\omega_0 = \exp(\xi_0)$

$$G(\xi, \varphi) = A \exp\left(-\frac{(\xi - \xi_0)^2}{2\sigma_\xi^2}\right) \exp\left(-\frac{(\varphi - \varphi_0)^2}{2\sigma_\varphi^2}\right) \quad (10)$$

where $A$ is a normalisation constant and $(\xi, \varphi)$ are the log-polar frequency coordinates, with $\xi = \log|\boldsymbol{\omega}|$ and $\varphi = \tan^{-1}(\omega_x, w_y)$. Gabor responses are grouped into a single complex vector with $n = 5 \times 6$ values per sampling grid, which is used as identity model. Matching between two images is done using the magnitude of complex values. Prior to matching with magnitude vectors, they are normalised to a probability distribution function by dividing each element of the vector by the sum of all vector elements, and matching is done using the $\chi^2$ distance [64]. In our previous works [11, 15], we accounted for rotation by shifting the sampling grid of the query image in counter- and clock-wise directions, and selecting the lowest matching distance. However, we observed that no significant improvement is achieved, allowing computational savings by removing such step. Therefore, rotation compensation will not be performed in this paper.

**Table 3** Configurations of the sampling grid with the databases used in this paper

| | BioSec | CASIA | IITD | MobBIO | UBIRIS |
|---|---|---|---|---|---|
| | | | Dense sampling | | |
| distance ($d_1$) | 30 | 30 | 30 | 16 | 16 |
| points | 13 × 19 = 247 | 9 × 11 = 99 | 9 × 13 = 117 | 9 × 13 = 117 | 19 × 23 = 437 |
| | | | Coarse sampling | | |
| distance ($d_2$) | 60 | 60 | 60 | 32 | 32 |
| points | 7 × 9 = 63 | 5 × 5 = 25 | 5 × 7 = 35 | 5 × 7 = 35 | 9 × 11 = 99 |

## 4 Baseline iris matchers

We conduct matching experiments of iris texture using four different systems based on 1D LG filters [19], local intensity variations in iris textures (CR) [20], DCT [21] and cumulative-sum-based grey change analysis (KO) [22]. We have used the LG implementation of Libor Masek [19], whereas the other three algorithms are from the University of Salzburg Iris Toolkit (USIT) software package [65]. In the four algorithms, the iris region is first unwrapped to a normalised rectangle using the Daugman's rubber sheet model [31]. Normalisation produces a 2D array (of 20 × 240, height × width, in the LG and 64 × 512 in the other three algorithms), with horizontal dimensions of angular resolution and vertical dimensions of radial resolution. Feature encoding is implemented according to the different extraction methods employed. The CR algorithm employs a template of integer values, which is matched via square differences, whereas the other three algorithms employ binary iris codes, which are matched using the Hamming distance.

## 5 Databases and protocol

As experimental dataset, we use data from the following six databases in our experiments: BioSec baseline [23], CASIA-Iris Interval v3 [24], IIT Delhi v1.0 [25], MobBIO [26], UBIRIS v2 [27] and Notre Dame ND-IRIS-0405 [28]. A summary of the used subset of these databases is given in Table 4. There are four databases acquired with NIR illumination, and two databases with visible (VW) light. All NIR databases use a close-up iris sensor, and they are mostly composed of good quality, frontal view images. MobBIO database has been captured with a Tablet personal computer, with two different lightning conditions, variable eye orientation and occlusion levels (distance to the camera was kept constant, however). UBIRIS v2 has been acquired with a digital camera, with the first session performed under controlled conditions, simulating an enrollment stage. The second session, on the other hand, was captured under a ‘real-world’ setup, with natural luminosity, heterogeneity in reflections and contrast, defocus, occlusions and off-angle images. Moreover, images of UBIRIS v2 have been captured from various distances.

**Table 4** Databases used and experimental protocol

| Database | Subjects | Eyes | Sessions | Images | Image size | Sensor | Lightning | Other information | Matching scores | |
|---|---|---|---|---|---|---|---|---|---|---|
| | | | | | | | | | Genuine | Impostor |
| BioSec [23] | 75 | 150 | 2 | 1200 | 480 × 640 | LG EOU3000 | NIR | four images/eye/ session, indoor | 2400 | 22 350 |
| CASIA Interval v3 [24] | 249 | 396 | 2 | 2655 | 280 × 320 | close-up camera | NIR | images per eye/ session not constant, indoor | 9018 | 146 667 |
| IIT Delhi v1.0 [25] | 224 | 448 | 1 | 2240 | 240 × 320 | JIRIS JPC1000 | NIR | five images/eye, indoor | 4800 | 200 256 |
| MobBIO [26] | 100 | 200 | 1 | 800 | 200 × 240 | Asus TE300T | visible | four images/eye; variable light, orientation, occlusion | 1200 | 39 800 |
| UBIRIS v2 [27] | 104 | 208 | 2 | 2250 | 300 × 400 | Nikon E5700 | visible | ≤15 images/eye/ session; variable light, orientation, occlusion | 15 750 | 22 350 |
| ND-IRIS-0405 [28] | 30 | – | – | 837 | 480 × 640 | LG 2200 | NIR | no user information available, indoor | – | – |

The six databases have been annotated manually by an operator [29], meaning that the radius and centre of the pupil and sclera circles are available, which are used as input for the experiments. Similarly, the eyelids are modelled as circles, which are used to build the noise mask of the iris matchers. Examples of annotated images are shown in Fig. 6. This segmentation groundtruth has been made available for the research community under the name of IRISSEG Database [29], and can be freely downloaded (http://www.islab.hh.se/mediawiki/index.php/Iris_Segmentation_Groundtruth and http://www.wavelab.at/sources).

We carry out verification experiments in this paper. We consider each eye as a different user (the number of available eyes per database is shown in Table 4). Genuine matches are as follows. When the database has been acquired in two sessions, we compare all images of the first session with all images of the second session. Otherwise, we match all images of a user among them, avoiding symmetric matches. Concerning impostor experiments, the first image of a user is used as enrolment sample, and it is matched with the second image of the remaining users. When the database has been acquired in two sessions, the enrolment sample is selected from the first session, and query samples are selected from the second session. The exact number of matching scores per database is given in Table 4. For the Notre Dame ND-IRIS-0405 database, there are few subjects which have been manually segmented. For this reason, this database has not been used in the verification experiments (note the ‘–’ in Table 4).

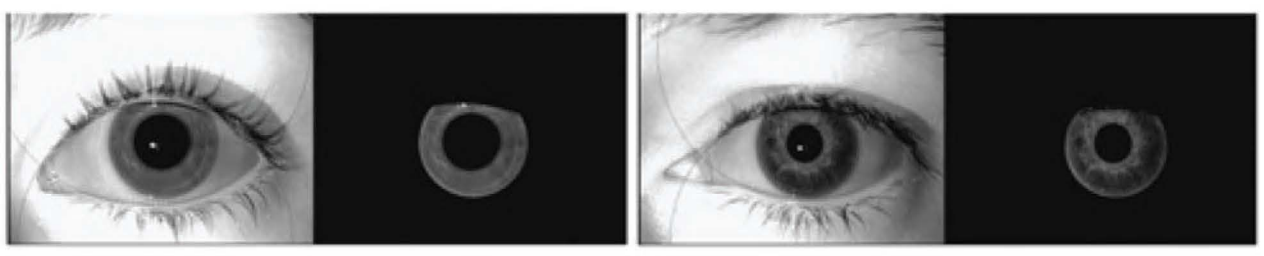

**Fig. 6** *Example of images of the BioSec database with the annotated circles modelling iris boundaries and eyelids*

Some fusion experiments are also done between different matchers. The fused distance is computed as the mean value of the distances because of the individual matchers, which are first normalised to be similarity scores in the [0, 1] range using tanh-estimators as $s' = (1/2)\{\tanh(0.01((s - \mu_s)/\sigma_s)) + 1\}$. Here, $s$ is the raw similarity score, $s'$ denotes the normalised similarity score and $\mu_s$ and $\sigma_s$ are, respectively, the estimated mean and standard deviation of the genuine score distribution [66].

## 6 Eye detection results

### 6.1 Setup

In Fig. 7, we give the performance of our eye detection system on the six databases used in this paper. The histograms of average width of the edge transitions (using the algorithm of Section 2.3) are shown in Fig. 8, and the histograms of pupil and sclera radii of each database (as given by the groundtruth) are shown in Fig. 9.

The eye detection system is evaluated under the following four scenarios, which have a different degree of adaptability to the width of the edge transitions of the input image. A ranked performance of the scenario 4 is also given in Table 5, with results of the other three scenarios also shown for comparative purposes:

(1) not including eyelashes removal, and parameter $\sigma$ of the first-order Gaussian-derivative filter (7) fixed to $\sigma = 7/6$ for all input images; this corresponds to a CDF transition of width $T = 7$ (9), which is the value used in our previous paper [15];
(2) including eyelashes removal ($L = 7$, based on [18]), and $\sigma = 7/6$;
(3) not including eyelashes removal, and $\sigma = T'/6$, with $T'$ being the edge transition width of the input image given by Section 2.3; and
(4) including eyelashes removal (with $L$ adapted to the input image as $L = T'$), and $\sigma = T'/6$.

The symmetry filter used is designed to cover 75% of the shortest image side. This is to ensure that it captures the different sizes of the eyes present in the databases because of variations in the distance to the sensor. Detection

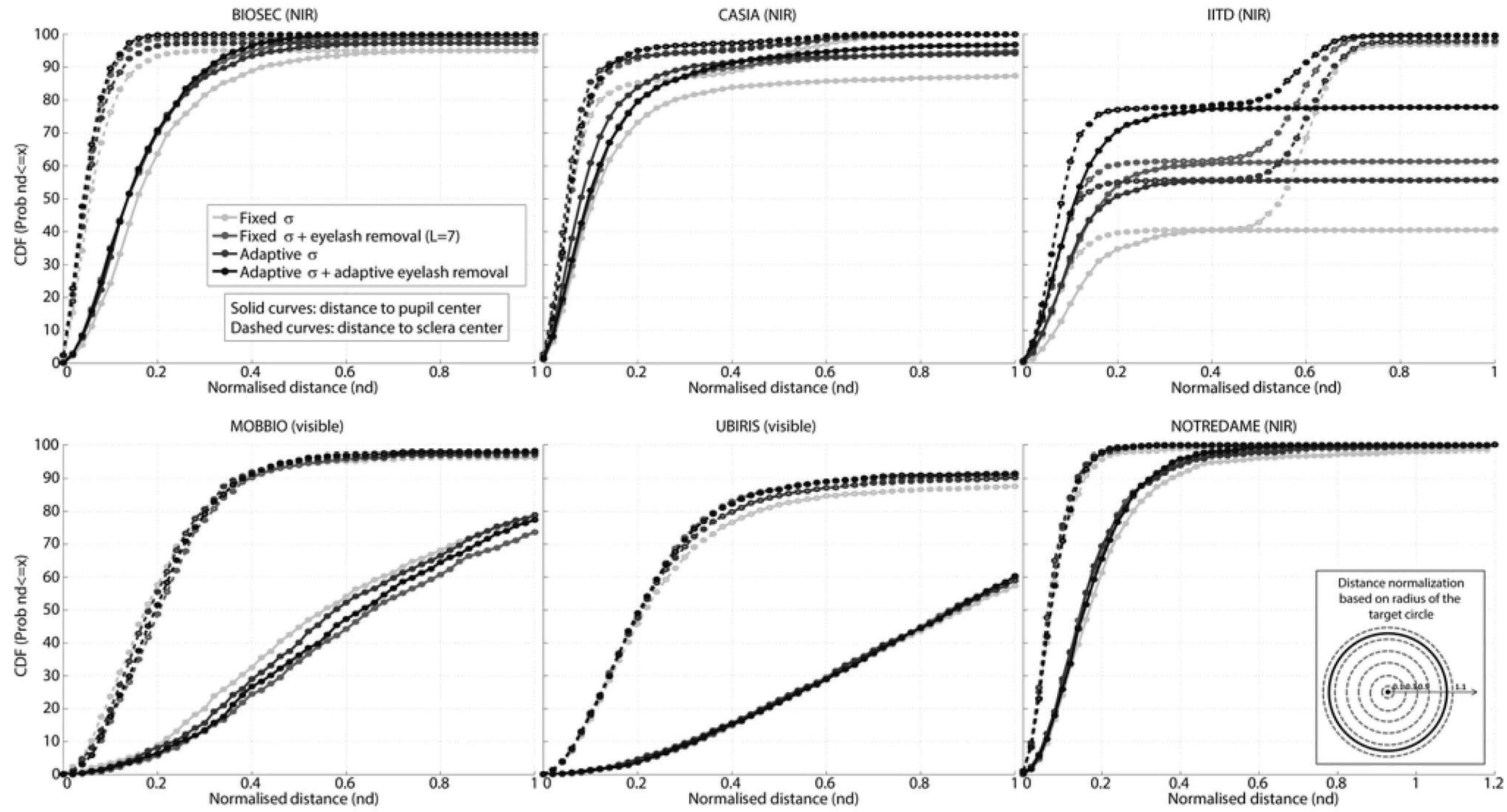

**Fig. 7** *Performance of automatic eye detection*

Inner sub-figure (bottom right): relative distance in terms of the radius of the target circle. The distance is normalised by the radius of the annotated circle for size and dilation invariance accuracy is evaluated by the distance of the detected eye centre with respect to the annotated pupil and sclera centres [67]. Distances are normalised by the radius of the annotated circles for size and dilation invariance, as shown in the inner sub-figure of Fig. 7 (bottom right). This way, a normalised distance *nd* lower than 1 means that the detected point is inside the circle, and the opposite if $nd > 1$. Moreover, since the sclera radius is always lower than the pupil radius, the normalised distance with respect to the sclera centre will be smaller than with respect to the pupil centre, as can be observed in Fig. 7.

### 6.2 Frequency estimation

An analysis of the groundtruth histograms (Fig. 9) reveals that all databases acquired with NIR illumination have approximately the same range of pupil and sclera radius, despite the use of sensors from different manufacturers. This is consistent with the fact that acquisition with this type of close-up sensors is done in a controlled manner, with the user always positioning the eye approximately at the same distance from the sensor. Only the Notre Dame database has slightly higher values. This explains that the histogram of edge transitions of this database is the right-most one (Fig. 8), meaning wider edge transitions, since the eye is closer to the camera. The case of IITD is particular, since its histogram of edge transitions is the left-most one (meaning shorter transitions on average). Images of IITD, however, shows very crisp details of the iris texture and surrounding regions (see Fig. 5), which may result in many short edge transitions because of iris texture details that are captured by the frequency estimation algorithm. Having iris images with more clear details seems a consistent explanation, since the best verification rates are obtained with IITD (as will be seen later in this section); however, this should be confirmed by additional studies. Concerning databases acquired with visible illumination, the pupil and sclera radii have lower values on average, meaning that the eye appears smaller in the image (see Fig. 5). With MobBIO, the range of radii is small, since it was acquired with constant distance to the camera [26]. In addition, the histogram of edge transitions in Fig. 8 appears towards the left (shorter transitions because of smaller eyes). UBIRIS, on the other hand, was intentionally acquired from various distances [27]. This is reflected in the spread of the histogram of sclera radii, which is wider than any other database.

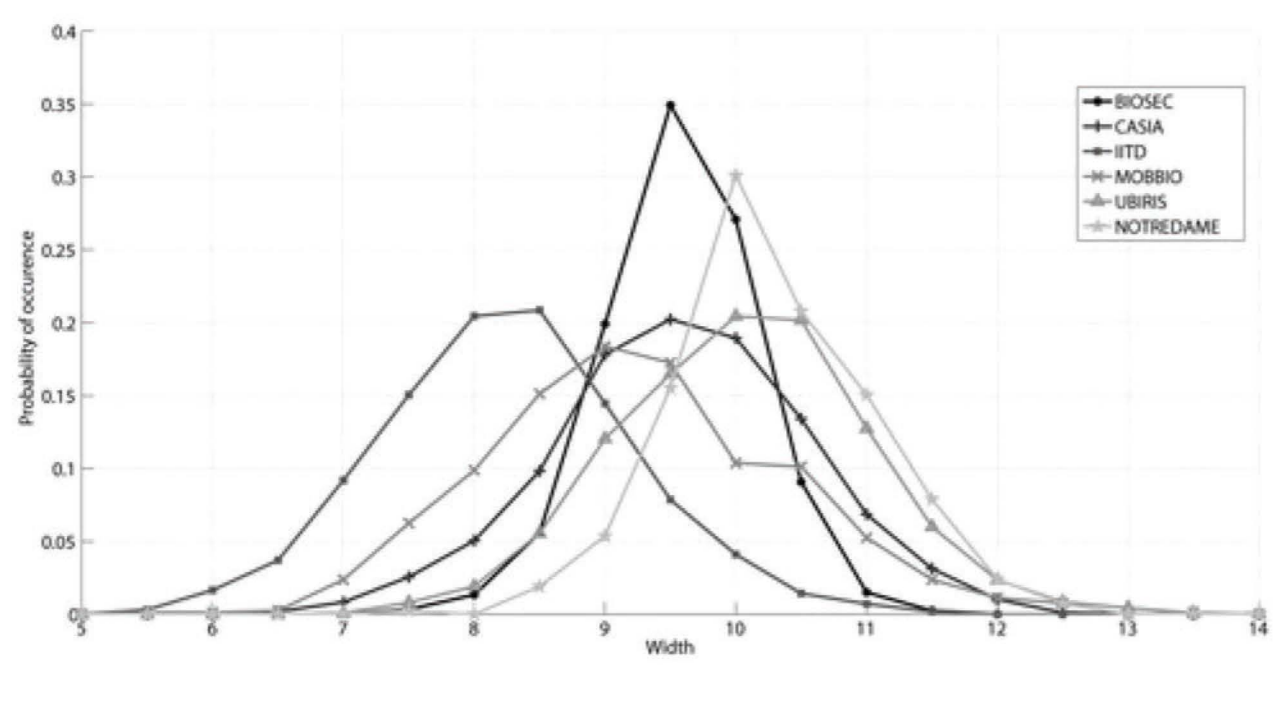

**Fig. 8** *Histograms of average width of edge transitions in iris images using the algorithm of Section 2.3*

### 6.3 Eye detection

Regarding the performance of the eye detection system (Fig. 7), databases acquired with NIR illumination show, in general, higher accuracy. These are databases that are acquired in a more constrained manner, controlling the illumination and the positioning of the person being captured [36]. Moreover, NIR images show a more detailed iris texture because of its higher reflectivity in this range

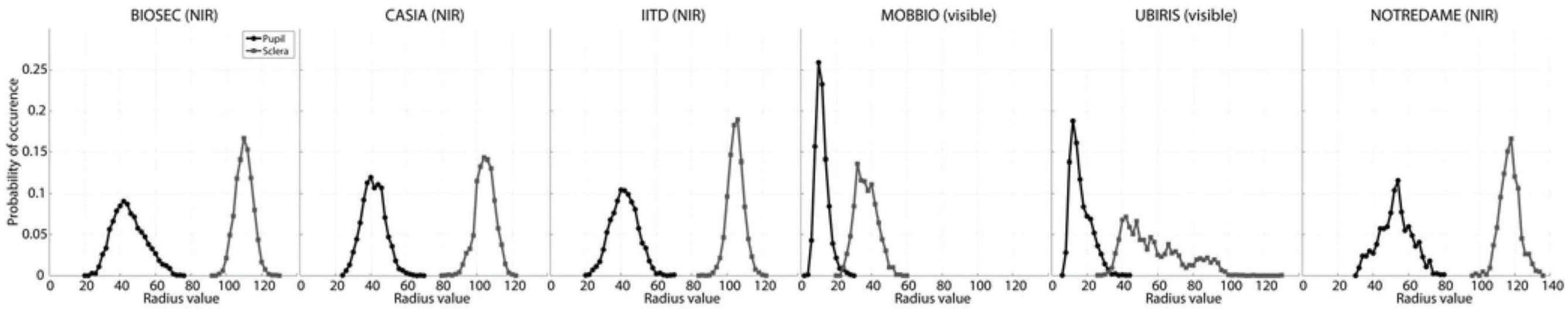

**Fig. 9** *Histograms of pupil and sclera radius of the databases used, as given by the groundtruth [29]*

[3]. This results in sharper iris boundaries, which are the features used by our eye detection algorithm. The detected eye centre with NIR databases usually falls within the pupil, that is, the solid curves in Fig. 7 reaches more than 95% for $nd \leq 1$. Moreover, for most images, the detected point is relatively close to the pupil centre ($nd \leq 0.3$ for ∼ 90% of the database). The only exception to this is (again) IITD. An explanation is that the sclera circle in this database is, in many cases, as big as the image itself (height of IITD images is 240 pixels, see Table 4, and half of the database has a sclera radius higher than 105 pixels, see Fig. 9). Since we are using a symmetry filter that covers 75% of the shortest image side (corresponding to a radius of 90 with IITD), there are a great amount of images in which the filter is not able to cover the (outer) sclera circle, relying on the (inner) pupil circle only.

With the databases acquired in visible range, eye detection is not so accurate: the detected eye centre falls within the pupil in about 80%/60% of the images (MobBIO/UBIRIS, respectively). With MobBIO, however, it is worth noting that the detected point is within the sclera for nearly the whole database (with UBIRIS, this happens in 90% of the images). This is a good result considering the more adverse conditions in which these databases have been acquired [26, 27]. Moreover, it should be considered that in these two databases, the eye appears smaller in the image, meaning that a displacement of few pixels in the detected eye position has higher impact in the normalised distance. Some examples of eye detection in images from MobBIO are given in Fig. 10. The three examples where detection occur in the vicinity of the eye centre (top) show cases of off-angle image (left), occluded eye (centre) and reflections because of glasses (right). The images below are examples of unsuccessful detection. Further examination of the first case (left) reveal that the inner (pupil) iris boundary is hardly visible, and as a result the filter response is weakened. The same happens in the second case (centre) because of occlusion. The third example (right) shows a maxima in the region of interest, but a stronger maxima occur because of curve-shaped boundaries given by the glasses of the contributor. Some other examples with images from UBIRIS are also given in Fig. 10. The first case of successful detection (top left) shows an image with glasses and uneven illumination. The second (centre) is an off-angle image, and the third (right) is an example with very low contrast. Concerning the examples of unsuccessful detection (bottom), in the first case (left) there is a stronger maxima because of curve-shaped boundaries of glasses. In the second case (centre), the eye is hardly visible because of occlusion, whereas the third (right) is an example of extreme off-angle combined with low contrast. As a result, in the latter image, the circular shape of the eyelids produces a stronger maxima.

Considering the four scenarios defined at the beginning of this section, it can be seen in Fig. 7 that in all cases, except MobBIO, the first scenario (grey curve) is the worst case. This is the scenario without eyelashes removal and with parameter $\sigma$ constant, meaning that including any preprocessing in the form of eyelash removal and/or adaptation to the width of the image edges is beneficial. The fourth scenario (black curve), when all preprocessing is adaptive to the estimated image frequency, is the best scenario with BioSec, CASIA and IITD, and it is always on top with the other three databases (except for the pupil curves of MobBIO). The case of IITD is representative. As explained before, the sclera circle is, in many cases, nearly as big as the image. It means that part of this circle will fall outside the image limits (see Fig. 5) because of displacements in eye positioning during acquisition. As a result, the eye detection system is more prone to errors with this database. A great amount of images, however, can be recovered by adding some kind of preprocessing, with the biggest benefit obtained by making adaptive both the eyelash removal step and the parameter $\sigma$ of the derivative filter (scenario 4). This extreme case shows the

**Table 5** Performance of automatic eye detection (ranking based on results of scenario 4)

| Database | Scenario 1 | | Scenario 2 | | Scenario 3 | | Scenario 4 | | Rank |
|---|---|---|---|---|---|---|---|---|---|
| | Pupil accuracy | Sclera accuracy | Pupil accuracy | Sclera accuracy | Pupil accuracy | Sclera accuracy | Pupil accuracy | Sclera accuracy | |
| BioSec (NIR) | 88.80 | 95 | 94.92 | 98.75 | 93.3 | 97.25 | 96.42 | 99.83 | 1 |
| ND (NIR) | 91.28 | 98.69 | 94.74 | 99.64 | 93.91 | 99.76 | 95.94 | 99.88 | 2 |
| CASIA (NIR) | 83.88 | 88.7 | 89.91 | 95.59 | 91.56 | 95.18 | 91.22 | 97.21 | 3 |
| IITD (NIR) | 40.36 | 40.63 | 60.72 | 61.65 | 55.36 | 55.71 | 77.32 | 78.39 | 4 |
| MobBIO (VW) | 32.5 | 91.13 | 24.5 | 90.25 | 29 | 91.88 | 26.75 | 91.25 | 5 |
| UBIRIS (VW) | 15.42 | 76.58 | 14.93 | 82.93 | 15.64 | 79.78 | 15.38 | 82.22 | 6 |

Results correspond to a detection error equal or <40% ($nd = 0.4$ in the $x$-axis of Fig. 7). The ranking is done based on accuracy with respect to the pupil centre.

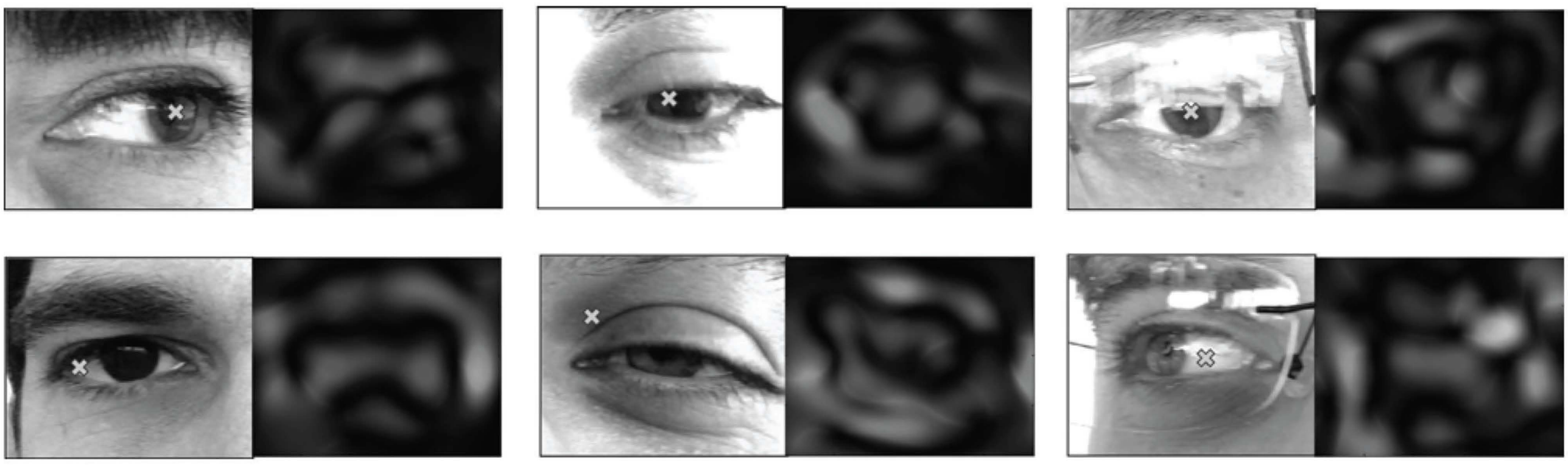

**Fig. 10** *Examples of eye detection with MobBIO*

*Top:* Successful localisation in the vicinity of the eye centre
*Bottom:* Unsuccessful localisation. Image $I_{20}$ is also given. The detected eye centre is marked with a prominent red cross benefits of the two new steps added to our eye detection system.

## 7 Recognition results

### 7.1 Setup

Next, we report verification results using the periocular system of Section 3 (Table 6). Results are given in the following two situations: (i) using as input the groundtruth pupil centre ('manual eye detection') and (ii) using the detected eye position given by our detection system ('automatic eye detection'). The latter are done with the eye detection system working under the fourth scenario defined in Section 6.1. Owing to different image sizes, Gabor filter wavelengths of the periocular system span the range 4–16 with MobBIO and UBIRIS (Fig. 11), and 16–60 with the other databases. For each database, this covers approximately the range of pupil radius of all its images, as given by the groundtruth (Fig. 9). Configuration of the sampling grid, including the number of sampling points per database, is given in Fig. 5 and Table 3. We consider two cases with the periocular system: (a) using the original iris images (Section 7.2) and (b) resizing the iris images to have a constant (average) sclera radius (Section 7.3). Finally, iris verification results and its fusion with the periocular matcher are provided, respectively, in Tables 7 and 8, which are analysed in Section 7.4.

**Table 6** Verification results in terms of EER (periocular system)

| Database | Periocular system | | | |
|---|---|---|---|---|
| | Manual eye detection | | Automatic eye detection | |
| | *d*1 | *d*2 | *d*1 | *d*2 |
| | Original image size | | | |
| BioSec (NIR) | 10.69 | 10.77 | 10.18 | 10.65 |
| CASIA (NIR) | 14.53 | 14.81 | 17.06 | 16.45 |
| IITD (NIR) | 2.5 | 2.67 | 10.62 | 10.76 |
| MobBIO (VW) | 12.65 | 15.16 | 14.31 | 15.15 |
| UBIRIS (VW) | 41.72 | 36.15 | 45.27 | 44.59 |
| | Resized images | | | |
| BioSec (NIR) | 10.56 | 10.91 | 10.24 | 10.24 |
| CASIA (NIR) | 15.55 | 15.4 | 17.71 | 17.07 |
| IITD (NIR) | 2.85 | 3.04 | 10.99 | 11.06 |
| MobBIO (VW) | 12.32 | 13.96 | 14.9 | 15.15 |
| UBIRIS (VW) | 24.81 | 24.4 | 35.49 | 35.44 |

### 7.2 Periocular recognition

As can be observed in Table 6 (top), results with automatic eye detection shows some degradation with respect to using groundtruth in databases where eye detection is less accurate (IITD, MobBIO and UBIRIS, according to Table 5). It is relevant however that, with MobBIO, no reduction in accuracy is observed with the coarse grid (compare results of '*d*2' columns). Recall that, with this database, the detected eye centre is within the sclera circle for nearly all images (Fig. 7). An explanation can be that since the dense grid has more points which are closer to each other, it is more sensitive to spatial displacements given by errors in the detection of the eye. This result is encouraging, since it shows that the periocular recognition system is able to cope with certain degree of error in estimating the eye centre when an appropriate grid configuration is chosen.

The degradation with IITD is specially significant, given its much worst results in eye detection observed in Fig. 7. It is also worth noting the very low EER of IITD with manual eye marking, which could be explained by the very crisp details observed in the iris texture and surrounding regions of the image, as mentioned before. Being able to improve the results in automatic eye detection with this database would result in a very powerful recognition system using our algorithm. On the other hand, UBIRIS shows a very high EER, which will be analysed in the following section. It is also worth noting the degradation observed in CASIA, despite the good results in eye detection with this database.

With respect to the use of dense or coarse grids, there are no appreciable differences in performance, either with manual or automatic eye detection (compare columns 2–3 and 3–4 in Table 6). This is good, considering that the dense grid has four times more points (Table 3). The only appreciable reduction in performance with a coarse grid is observed in MobBIO (apart from UBIRIS), which could be attributed either to the smaller size of the eye (Fig. 9) or the more adverse acquisition conditions in visible range, showing more variability in illumination, eye orientation and occlusion [26]. This should need additional experiments, however.

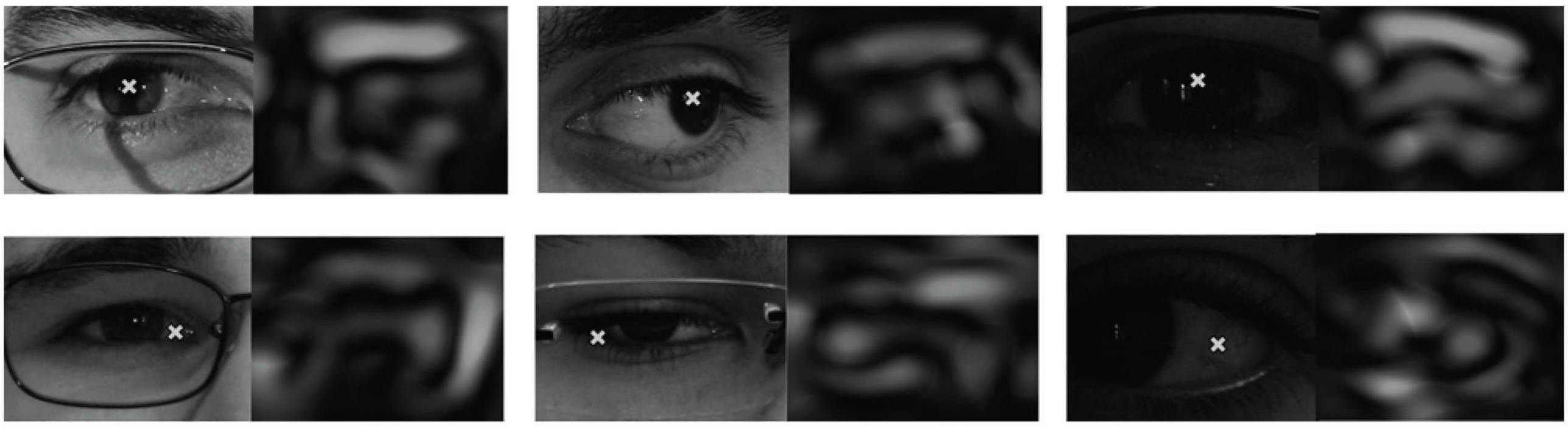

**Fig. 11** *Examples of eye detection with UBIRIS*

*Top:* Successful localisation in the vicinity of the eye centre
*Bottom:* Unsuccessful localisation. Image $I_{20}$ is also given. The detected eye centre is marked with a prominent red cross

### *7.3 Image resize*

From Table 6 (top), we observe a very poor performance in UBIRIS using the original iris images (EER of 36% or more). Our assumption is that apart from being the database acquired with the most adverse perturbations in terms of illumination, off-angle, occlusions etc. (Section 5), there is also a wide variability in eye resolution (see Fig. 9) because of acquisition at different distances. As a result, the points of the grid used by our periocular algorithm (which is of constant dimensions) are not capturing consistently the same region of the image (observe Fig. 12, top).

Motivated by this fact, we have conducted verification experiments where all images of the database have been resized via bicubic interpolation to have the same sclera radius. For each database, we choose as target radius the average sclera radius of the whole database, as given by the groundtruth. Verification results after this procedure are given in the bottom part of Table 6. As it can be observed, EER with UBIRIS is reduced significantly with this strategy. It is also of relevance that for the other databases, there is no substantial change in performance after images have been resized. This means that the periocular recognition is able to cope with small changes in the scale (size) of the eye. On the other hand, the performance with UBIRIS after image

**Table 7** Verification results in terms of EER (iris matchers)

| Database | Iris systems | | | | Periocular best case |
|---|---|---|---|---|---|
| | LG | CR | DCT | KO | |
| BioSec (NIR) | 1.12 | 12.93 | 2.31 | 10.64 | 10.56 |
| CASIA (NIR) | 0.67 | 8.85 | 1.73 | 13.44 | 14.53 |
| IITD (NIR) | 0.59 | 3.55 | 0.96 | 2.74 | 2.5 |
| MobBIO (VW) | 18.81 | 20.56 | 31.1 | 22.81 | 12.32 |
| UBIRIS (VW) | 35.61 | 37.87 | 47.46 | 34.94 | 24.4 |

Performance of the periocular system (best case with manual eye detection) is also shown for comparison purposes.

**Table 8** Verification results in terms of EER (fusion or periocular and iris systems)

| Database | Fusion: periocular ($d2$) + iris | | | | | | | |
|---|---|---|---|---|---|---|---|---|
| | Manual eye detection | | | | Automatic eye detection | | | |
| | Original image size | | | | | | | |
| | LG | CR | DCT | KO | LG | CR | DCT | KO |
| BioSec (NIR) | 2.16 | 7.61 (−29.34%) | 4.49 | 8.36 (−21.43%) | 1.96 | 7.68 (−27.89%) | 4.02 | 7.97 (−25.09%) |
| CASIA (NIR) | 2.38 | 8 (−9.6%) | 5.67 | 9.89 (−26.41%) | 2.5 | 8.37 (−5.42%) | 4.9 | 11.55 (−14.06%) |
| IITD (NIR) | 1.2 | 1.82 (−31.84%) | 1.63 | 1.91 (−28.46%) | 1.99 | 6.93 | 6.28 | 4.21 |
| MobBIO (VW) | 11.75 (−22.49%) | 12.13 (−19.99%) | 15.99 | 14.72 (−2.9%) | 12.34 (−18.55%) | 12.83 (−15.31%) | 17.4 | 15.96 |
| UBIRIS (VW) | 29.49 (−17.19%) | 33.19 (−8.19%) | 38.32 | 31.02 (−11.22%) | 35.06 (−1.54%) | 37.77 (−0.26%) | 43.63 (−2.15%) | 35.4 |
| | Resized image | | | | | | | |
| | LG | CR | DCT | KO | LG | CR | DCT | KO |
| BioSec (NIR) | 2.11 | 7.99 (−26.76%) | 4.61 | 8.47 (−20.39%) | 1.98 | 7.54 (−26.37%) | 3.86 | 8.18 (−20.12%) |
| CASIA (NIR) | 2.54 | 8.45 (−4.52%) | 6.03 | 10.13 (−24.63%) | 2.46 | 8.63 (−2.49%) | 5.01 | 11.8 (−12.2%) |
| IITD (NIR) | 1.47 | 2.24 (−26.32%) | 2.03 | 2.09 (−23.72%) | 2.07 | 6.89 | 5.87 | 4.26 |
| MobBIO (VW) | 11 (−21.2%) | 11.81 (−15.4%) | 14.6 | 14.19 | 11.68 (−22.90%) | 12.75 (−15.84%) | 16.29 | 15.12 (−0.2%) |
| UBIRIS (VW) | 22.41 (−8.16%) | 25.47 | 30.44 | 24.3 (−0.41%) | 28.05 (−20.85%) | 30.94 (−12.7%) | 36.19 | 29.42 (−15.8%) |

The relative EER variation with respect to the best individual system is given in brackets (only when there is performance improvement).

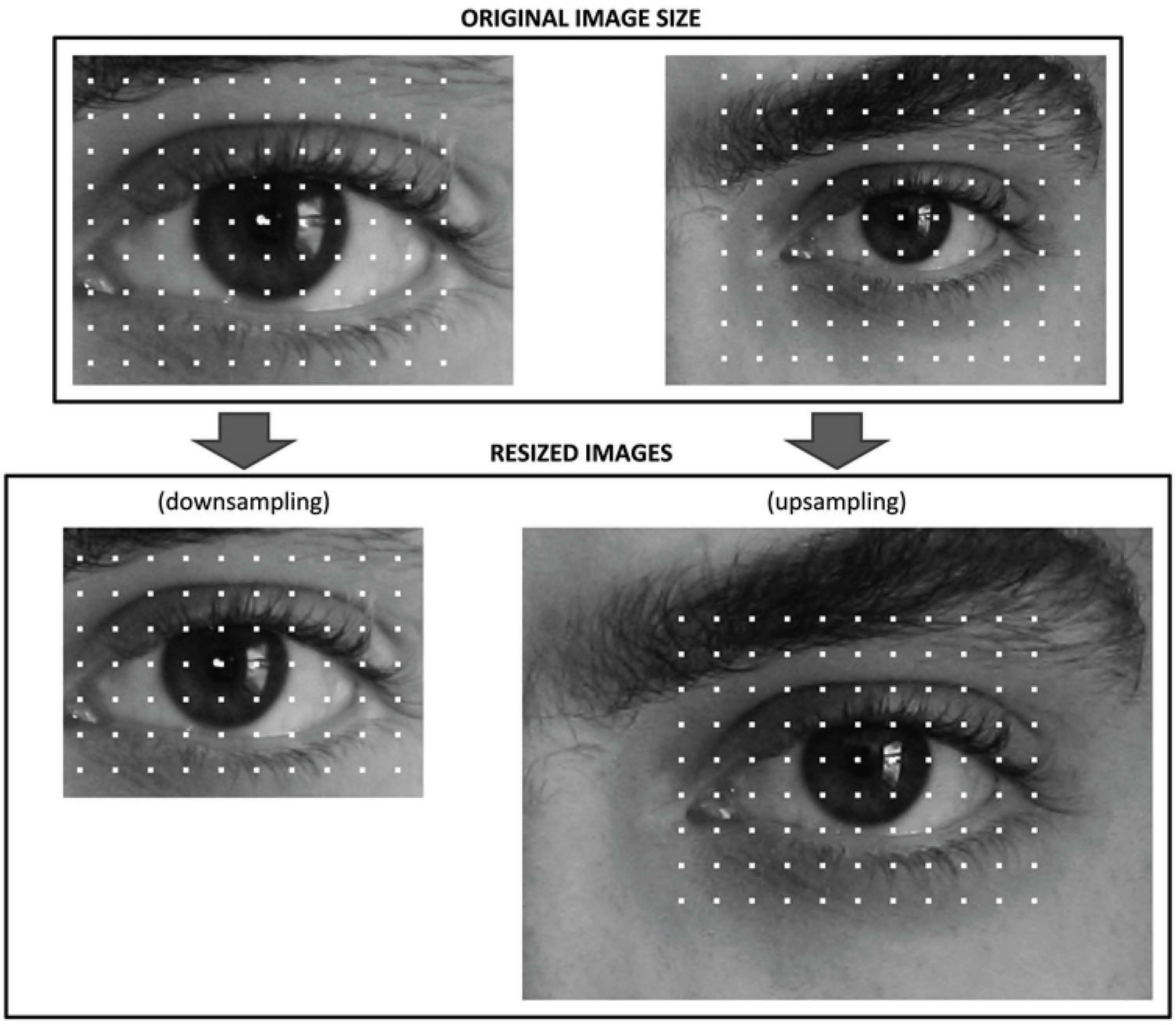

**Fig. 12** *Top: Example of grid positioning in two images from the same user having different eye resolution. Bottom: images resized to have the same sclera radius. Distance between sampling points is 32. Images are from UBIRIS*

resizing is still much worst than the other databases, which could be attributed to the remaining perturbations present in this database (lightning changes, off-angle etc.)

### *7.4 Fusion with the iris modality*

Results of the iris matcher and fusion with the periocular system are shown in Tables 7 and 8. Concerning the iris matchers, it is observed that their performance is, in general, much better than the periocular matcher with NIR databases (BioSec, CASIA, IITD). This is expected, since iris systems usually work better in NIR range [31], and it confirms our previous results using only BioSec and MobBIO databases [15]. The KO iris matcher shows similar performance than the periocular matcher with NIR databases, and the CR iris matcher performs worst with BioSec and IITD. Regarding absolute performance numbers, it is relevant that LG and DCT matchers have the best performance with NIR data, but DCT matcher does much worst than LG with VW data (see MobBIO and UBIRIS). Moreover, despite the worst performance of CR and KO with NIR data, their performance is comparable with LG with VW data. Although it is not the scope of this paper, these results seem to suggest that some of the iris features used are more suitable for NIR than for VW data, and vice-versa. It should be remarked however that the iris matchers have been executed without any image enhancement step (e.g. reflections removal or contrast equalisation), so the incorporation of such preprocessing, or the use of other software implementations, may lead to different results.

As regards to the fusion experiments (Table 8), the best iris matchers with NIR data (LG and DCT) do not result in performance improvement when fused with the periocular matcher. One reason could be the big difference in performance between iris and periocular found in these cases. On the other hand, fusion using the CR or KO matchers (which have comparable performance with the periocular matcher using NIR data, see Table 7) results in performance improvements of up to 32%. It is therefore obvious in our experiments that there is more benefit in fusing two modest matchers than in fusing one matcher that is already very good with another with a modest performance. This, however, should not be taken as a general statement. Other fusion rules different than the one employed here may lead to different results, specially if the supervisor is data quality and/or expert adaptive [66, 68], see, for example, the fingerprint experiments of [69].

It is of relevance that on the other hand, the periocular system works better than the iris matchers with VW databases (with UBIRIS, this happens after images are resized). One reason could be that the eye has a smaller size in these databases (see Fig. 9), so it is more difficult to extract reliable identity information from the (even smaller) iris texture. In such conditions, the periocular region is able to provide a rich source of identity data, as evidenced in our experiments. It is worth noting, however, that even in the adverse conditions of the VW databases, the iris texture is still able to complement the periocular system, as shown in the fusion results. Using the LG iris matcher, for example, EER improvements of 20% or more can be achieved by the fusion. On the other hand, despite the performance of CR and KO is comparable with LG, the fusion of CR/KO with the periocular system does not have similar performance improvements in many cases. Only CR with MobBIO shows consistent performance improvements of 15–20%. Finally, the DCT iris matcher (which performs the worst with VW data) does not provide any improvement when fused with the periocular system.

## 8 Conclusions

Periocular recognition has emerged as a promising trait for unconstrained biometrics [1–3], following demands for increased robustness of face or iris systems, with suggestions that it may be as discriminating by itself as the face as a whole [7, 8]. Periocular refers to the region in the immediate vicinity of the eye, including the eye, eyelids, lashes and eyebrows. It has shown a surprisingly high discrimination ability [1], and it can be easily obtained with existing setups for face and iris. A primary consequence is to drastically reduce the need of user cooperation. It is also available over a wide range of distances even when the iris texture cannot be reliably obtained (e.g. low resolution, blinking or closed eyes, off-angle poses and inappropriate illumination) or when portions of the face are occluded (e.g. close distances) [10]. Most face detection and recognition systems use a holistic approach, that is, they require a full-face image, so the performance is negatively affected in case of occlusion [5]. Moreover, the periocular region appears in iris images, so the fusion of the information present in the periocular region with the iris texture has a potential to improve the overall recognition [9].

In this paper, we propose an eye detection system based on symmetry filters. It is based on 2D separable symmetry filters tuned to detect circular symmetries, in such a way that we detect the eye centre with a few 1D convolutions. One advantage of this system is that it does not need training, in contrast to other previous works making use, for example, of Gabor features [13], Viola–Jones detectors of face sub-parts [37] or correlation filters [10]. This detection system is used as input to a periocular algorithm based on retinotopic sampling grids and Gabor analysis of the power spectrum. This framework is evaluated with six databases of iris data, four acquired with a close-up NIR camera, and two in VW light with a webcam and a digital camera. This paper expands our two previous studies where we presented the eye detection [15] and the periocular recognition [11] systems. In particular, the eye detection system includes two new additions. The first one is concerned with frequency estimation of the input iris image [16], which is used to dynamically adjust the derivative filters used to compute the image orientation field. The second is an eyelash removal step [18], which also made adaptive by using the estimated image frequency. The addition of these steps has demonstrated to improve the performance of our detection system. With NIR images, the detected eye centre is very close to the pupil centre (measured by groundtruth [29]). The two databases in visible range shows worst performance, because of more difficult acquisition conditions which include uneven illumination, off-angle eyes, occlusions and distance changes. In one of the visible databases, however, the detected point is within the sclera for nearly the whole database. This is enough for the recognition algorithm, since no reduction in performance is observed in this database with an appropriate configuration of the sampling grid.

As far as the periocular recognition system is concerned, it is shown to be robust to a certain degree of inaccuracy in detecting the eye centre, also being able to cope with small-scale changes in the eye image. Dimensions of the grid are constant for all images of each database, without adaptation to the size of the input eye, so the only requirement is the availability of the eye centre. We also test two different sampling grid configurations, with dense and coarse sampling. It is observed that the accuracy of the periocular matcher is not jeopardised by reducing the density of the sampling grid. This is good news for time saving purposes, considering that the dense grid has four times more less points. Lastly, we evaluate four iris matchers based on 1D LG filters [19], local intensity variations in iris textures (CR) [20], DCT [21] and cumulative-sum-based grey change analysis (KO) [22]. The performance of the iris matchers are, in general, much better than the periocular matcher with NIR data, and the opposite with VW data. However, there is not a general trend among all the iris matchers, suggesting that some iris of the iris features used are more suitable for NIR than for VW data, and vice-versa. Regarding fusion experiments, despite the poorer performance of the iris matchers with the VW databases, its fusion with the periocular system can provide an improved performance (of more than 20% with one of the iris matchers). This is remarkable given the smaller eye size on VW databases, resulting in an (even smaller) iris region. With NIR images, the fusion only improves if the iris and periocular matcher have comparable performance; however, for the best iris matchers (EER about 1–2% or less), there is no improvement by the fusion because of the big difference in performance with respect to the periocular matcher. This paper also presents a new groundtruth database of iris segmentation data [29]. The six databases used in this paper have manually annotated by an operator, with the annotation being made available to the research community.

Future work includes evaluating the reliability of the proposed eye detection system in full-face images. Another source of improvement will be the incorporation of a refinement stage, for example, by pixel-wise analysing the neighbourhood of the detected point [4] to achieve a more accurate estimation of the eye centre with visible images. We also plan to evaluate other existing periocular recognition algorithms [1]. Since we have observed that images in visible, in general, provide worst performance in our developments, we will also focus on this type of images, finding mechanisms to cope with more adverse images in terms of uneven lightning, reflections, off-angle, occlusions or de-focus.

## 9 Acknowledgments

F. A.-F. thanks the Swedish Research Council and the EU for funding his postdoctoral research. The authors acknowledge the CAISR programme of the Swedish Knowledge Foundation, the EU BBfor2 project and the EU COST Action IC1106.